\documentclass{article}

\usepackage[utf8]{inputenc}
\usepackage{color}
\usepackage{graphicx}
\usepackage{units}  % For \unit.
\usepackage{url}
\usepackage{coverpage}

\begin{document}

\title{Ego-Motion Sensor for Unmanned Aerial Vehicles Based on a Single-Board Computer}

\author{Ga\"el \'Ecorchard\footnote{Gaël Écorchard's work is supported by the SafeLog project funded by the European Union's Horizon 2020 research and innovation programme under grant agreement No.~688117.}, Adam Heinrich and Libor P\v{r}eu\v{c}il\footnote{Libor Přeučil's work is supported by the Technology Agency of the Czech Republic under Project TE01020197 Center for Applied Cybernetics.}\\
Czech Institute for Informatics, Robotics, and Cybernetics\\
Czech Technical University in Prague, Czech Republic\\
E-mail: gael.ecorchard@cvut.cz, https://www.ciirc.cvut.cz
%\thanks{$^{1}$ Gaël Écorchard's work is supported by ...}
}

\date{}

\published{\it Human-Centric Robotics, Proceedings of CLAWAR 2017: 20th International Conference on Climbing and Walking Robots and the Support Technologies for Mobile Machines, pp. 189--196. DOI: 10.1142/9789813231047\_0025}
\original{https://doi.org/10.1142/9789813231047\_0025}

\coverpage

\maketitle

\begin{abstract}
This paper describes the design and implementation of a ground-related odometry sensor suitable for micro aerial vehicles.
The sensor is based on a ground-facing camera and a single-board Linux-based embedded computer with a multimedia System on a Chip (SoC).
The SoC features a hardware video encoder which is used to estimate the optical flow online.
The optical flow is then used in combination with a distance sensor to estimate the vehicle's velocity.
The proposed sensor is compared to a similar existing solution and evaluated in both indoor and outdoor environments.
\end{abstract}

\leftskip18pt\rightskip\leftskip
    \par\vskip-12pt
%\keywords{visual odometry, optical flow, ego-motion}

\section{Introduction}

The ability to estimate the velocity is a fundamental task for the control of micro aerial vehicles (MAVs).
Such information can be provided, e.g.\ by the PX4Flow sensor\cite{Honegger_2013}, which is an optical flow sensor based on a microcontroller.
The sensor provides the linear velocities at a rate of \unit[400]{Hz} but does not provide changes in heading (yaw).
A similar solution is the already-discontinued ArduEye \cite{ArlArduFlow}.
The visual odometry method described by Kazik and Goktoganin \cite{Kazik_2011} is based on a ground-facing camera and an approach based on a Fourier-Mellin transform which recovers both rotation and translation instead of using the optical flow.

Optical flow sensors based on chips used in optical computer mice used to be quite popular especially in the hobby community, probably due to the good availability and low cost of these sensors, such as the ADNS-3080 chip\cite{ADNS-3080}.
One of the available solutions using such technology is a part of the ArduPilot system \cite{ArduPilot}, however, a rotation around the center of the sensor (yaw) can not be recovered and is said to confuse the sensor.
Some authors, e.g. Briod et al.\cite{briod2013optic} or Kim and Brambley \cite{Kim2007}, combine several such sensors and an inertial unit, thus removing the need for a separate distance sensor.

A common approach which enables the usage of otherwise CPU-intensive algorithms in real-time systems is their implementation in FPGA, cf. Krajnik et al.\cite{Krajnik_2012}.
The displacement of SURF features between consecutive frames is used to estimate the MAV's displacement.

Next section introduces some theoretical background related to optical-flow sensors.
Then, we present our implementation before presenting the real-world results and concluding.

\section{Theoretical Background}

\subsection{Optical Flow}

An optical flow is the displacement of pixel values in the image sequence induced by a movement of a camera or a scene observed by it.
Let $I(u,v,t)$ be an image function of the pixel position $(u,v)$ and time $t$.
The optical flow between two frames captured at times $t$ and $t+\Delta t$ can then be represented by the displacement $(\Delta u,\Delta v)$ and time difference $\Delta t$.

% An example optical flow visualized using vector field is depicted on figure~\ref{fig:flow-raspivid-example}.

% \begin{figure}[H]
%   \includegraphics[width=0.9\columnwidth]{drawings/optical_flow_example}
%   \caption{Optical flow induced by a object moving downwards\label{fig:flow-raspivid-example}}
% \end{figure}

Most approaches to optical flow estimation are based on a brightness
constancy constraint.
This constraint assumes that moving pixels keep the same brightness between consecutive frames.

The brightness constancy can be linearized using the Taylor approximation \cite{Wedel_StereoFlow}, which yields in:
\begin{equation}
  0 = I_{u}V_{u} + I_{v}V_{v} + I_{t},
  \label{eq:of_brightness_constancy_equation}
\end{equation}
where $I_{u}$, $I_{v}$ and $I_{t}$ are partial derivatives of the image function with respect to $u$, $v$ and $t$, respectively and $V_{u}$ and $V_{v}$ are the velocities of the optical flow, $V_{u} = \frac{\Delta u}{\Delta t}$ $V_{v} = \frac{\Delta v}{\Delta t}$.

As Equation~\ref{eq:of_brightness_constancy_equation} has two variables, it has an infinite number of solutions, this is known as the aperture problem.
This ambiguity means that another constraints have to be enforced, such as the spatial smoothness constraint.
The spatial smoothness constraint assumes that neighboring pixels belong to the same objects and therefore represent the same motion.

The block matching algorithm is one of the simplest methods to compute the optical flow.
For every pixel $(u,v)$ in the original image, the closest match $(u+\Delta u,v+\Delta v)$ in the subsequent image is found by minimizing the Sum of Absolute Differences (SAD).
The SAD value is computed by comparing a small (usually square) window $(M\times N)$ around the pixel:
\begin{equation}
  SAD = \sum_{i=-\frac{M}{2}}^{\frac{M}{2}}
          \sum_{j=-\frac{N}{2}}^{\frac{N}{2}}
            \left| I(u+\Delta u+i,v+\Delta v+j,t+\Delta t) - I(u+i,v+j,t) \right|
  \label{eq:of_sad}
\end{equation}

This method can be made faster by computing the flow only for a subset of image pixels instead of the full image matrix, producing only a sparse optical flow.
It can be well parallelized as the flow can be computed independently for each pixel.

\subsection{Camera Model}

% The camera model is simplified by the pinhole camera model represented Fig.~\ref{fig:pinhole-camera}
%
% \begin{figure}[!ht]
%   \centering
%   \includegraphics{figures/camera.pdf}
%   \caption{Pinhole camera model}
%   \label{fig:pinhole-camera}
% \end{figure}

The camera model is simplified by the pinhole camera model.
By supposing that the vertical position of the camera is approximately constant between two image frames, the displacement of a point in space $[\Delta X,\Delta Y,0]^{T}$ between two image frames is associated to the displacement of the point projected by a pinhole camera onto the image plane $[\Delta u,\Delta v, 0]^{T}$ in pixels, which can be computed as
\begin{equation}
  \Delta X = -\frac{s}{f}\Delta u Z,\;
  \Delta Y = -\frac{s}{f}\Delta v Z,
  \label{eq:px2m}
\end{equation}

\noindent where $f$ is the focal length, $s$ is the pixel size, $Z$ is the distance from the camera to the ground.

The ground distance $Z$ must be obtained from an external sensor, such as ultrasonic, laser, or barometric pressure sensors.

\subsection{Compensation of Angular Velocities}

It is necessary to compensate small rotations between consecutive frames which manifest as an optical flow in the image plane.
Assuming that the camera has been rotated around its $x$ and $y$ axes between two consecutive
frames, the displacements in the image plane induced by the rotations are
\begin{equation}
  \Delta x' = f\tan(\omega_{y}\Delta t),
  \Delta y' = f\tan(\omega_{x}\Delta t),
  \label{eq:angular_correction}
\end{equation}
\noindent where $\omega_{x}$ and  $\omega_{y}$ are the angular velocities which can be obtained from a gyroscope.

The displacements $\Delta x'$ and $\Delta y'$ have to be subtracted from the resulting optical flow in order to compensate for the angular motion.
The rotation around the optical axis ($z$-axis) does not have to be corrected as the induced optical flow is useful for the estimation of the vehicle's heading.
By approximating $\tan(x) \approx x$ for the small involved angles, Equation~\ref{eq:px2m} becomes then
\begin{equation}
  \Delta X = -(\frac{s}{f}u-f\omega_{y}\Delta t) Z,
  \Delta Y =-(\frac{s}{f}v-f\omega_{x}\Delta t) Z.
  \label{eq:px2m_corr}
\end{equation}

\section{Implementation}

The algorithm presented in the previous section was implemented on the Raspberry Pi 3 Single-Board Computer as a mixed CPU--GPU solution\cite{RaspeberryPiWebsite}.
The Raspberry Pi was chosen for its low cost and the ability to obtain the optical flow from its integrated hardware H.264 encoder, through its undocumented Coarse Motion Estimator (CME).

\begin{figure}[ht]
  \centering
  \includegraphics[width=0.5\columnwidth]{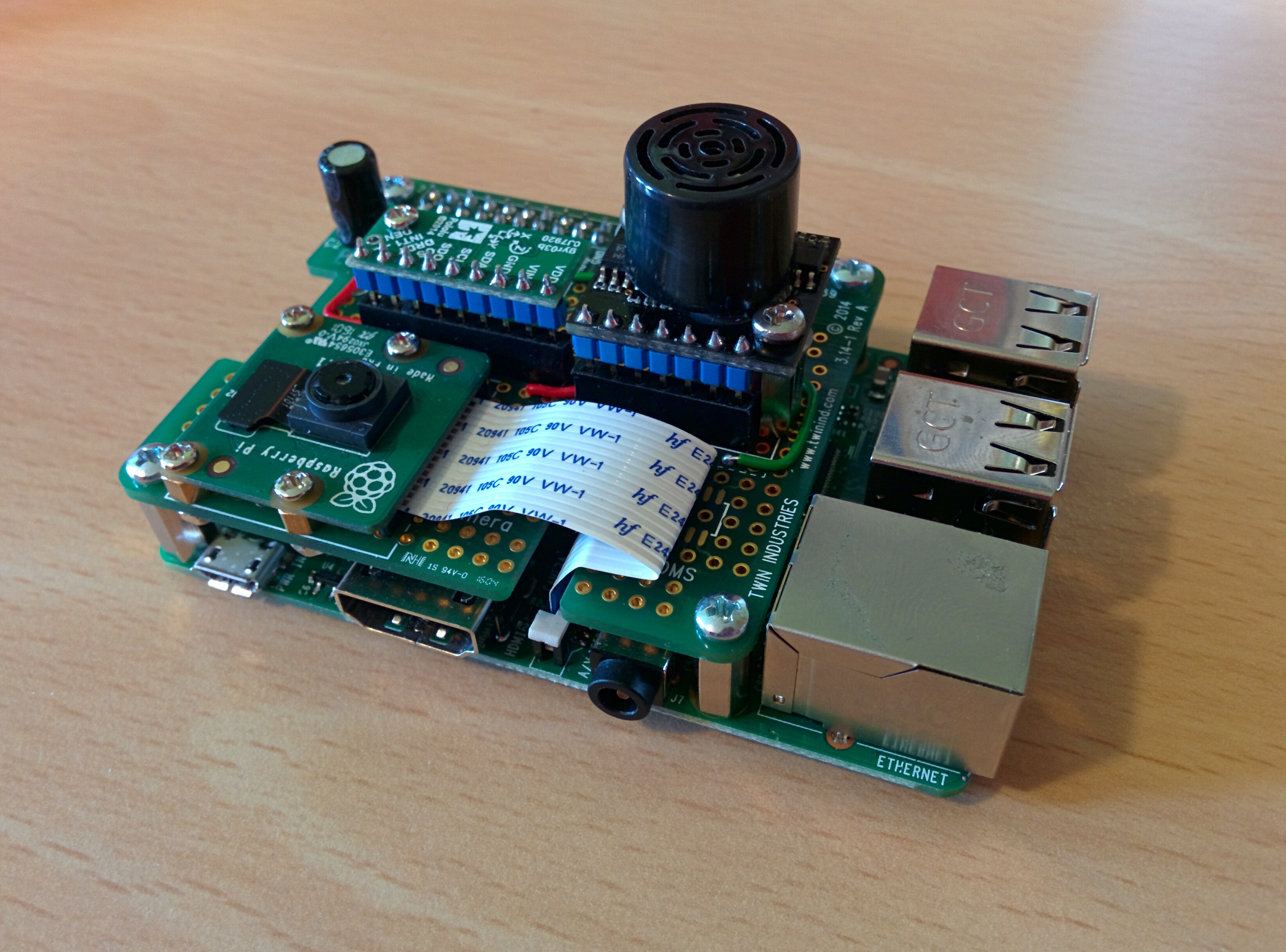}
  \caption{Raspberry Pi with camera and sensors}
  \label{fig:rpi_pcb}
\end{figure}

The other required hardware used in the current setup is the Raspberry Pi camera module v2, with a Sony IMX219 image sensor, used at a resolution of 1640 $\times$ 1232 at \unit[40]{fps} or 1280 $\times$ 720 at \unit[90]{fps}, the MaxBotix HRLV-EZ4 ultrasonic distance sensor, and the L3GD20H 3-axis MEMS digital gyroscope by STMicroelectonics.
The complete setup is presented in Fig.~\ref{fig:rpi_pcb}.

\begin{figure}[!ht]
  \centering
  \includegraphics[width=\columnwidth]{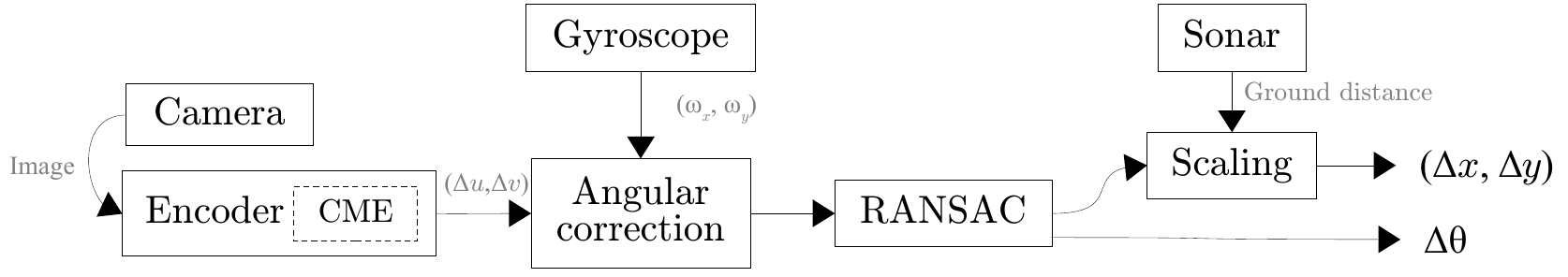}
  \caption{Pipeline of the designed solution}
  \label{fig:design_pipeline}
\end{figure}

The different steps of the algorithm are described in Fig.~\ref{fig:design_pipeline}.
The RANSAC algorithm computes the rigid transform while eliminating outliers.

The CME divides the image into 16$\times$16-pixel macroblocks and provides the motion vectors as two 8-bit values and the SAD as a 16-bit value per macroblock. Only the motion vectors are used by our algorithm.

Although the displacements can theoretically be in the range $\pm$127 pixels, a closer analysis with a rapidly moving video sequence showed that it is in fact in the range $\pm$64 pixels from the macroblock's center with a two-pixel resolution.

\begin{figure}[ht]
  \centering
  \includegraphics[width=0.7\columnwidth]{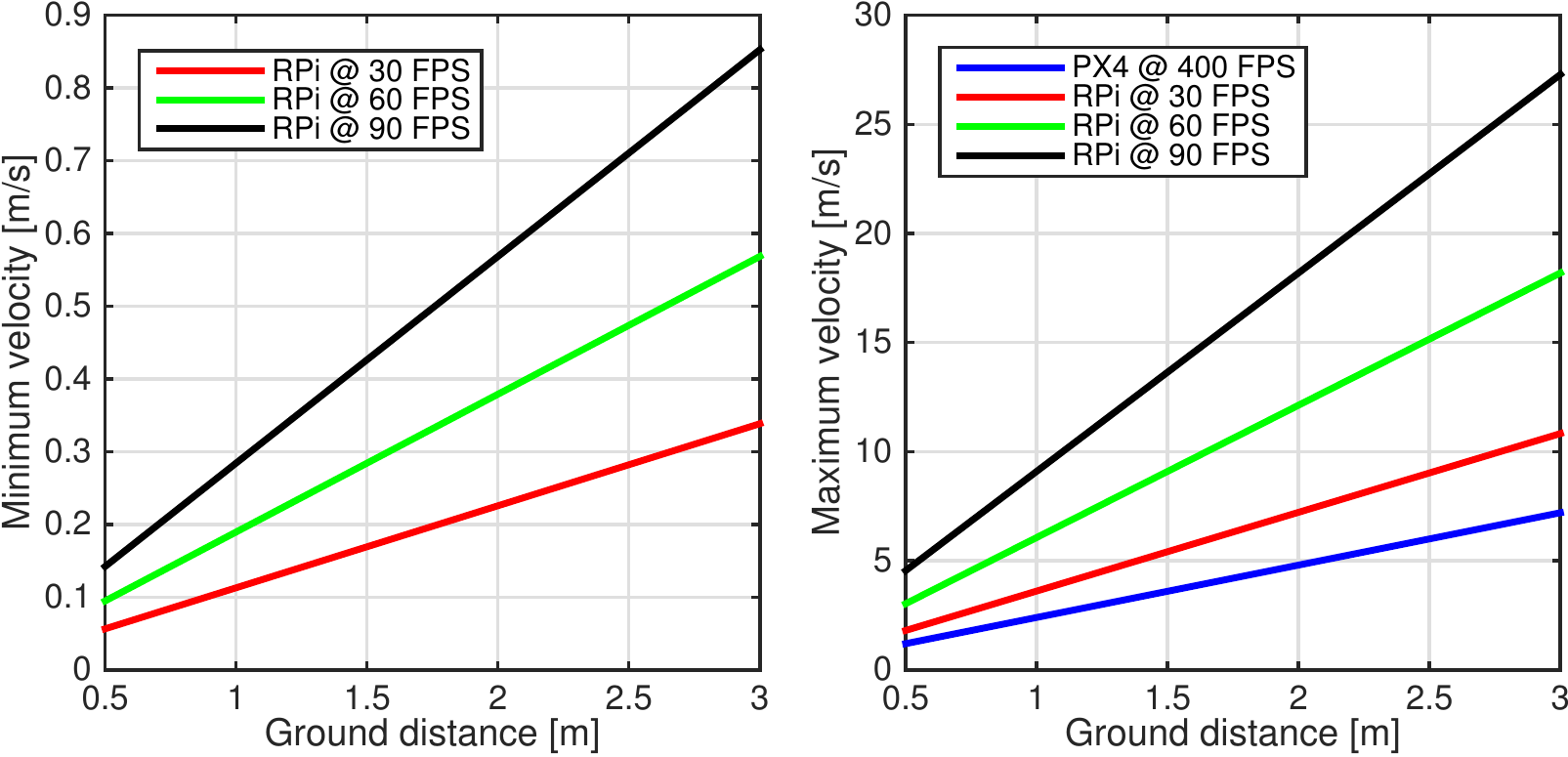}
  \caption{Theoretical velocity limits with respect to the ground distance}
  \label{fig:theoretical_min_max_velocity}
\end{figure}

With the camera parameters obtained by calibration and the range of the motion vector, one can compute the minimum and maximum theoretical detectable velocities.
These results are presented in Fig.~\ref{fig:theoretical_min_max_velocity} and compared to those of the PX4Flow. Although our system's frame rate is low when compared to the PX4Flow, it operates with a larger resolution of 480~$\times$~480, so its maximum theoretical detectable velocity is even higher.

\section{Results}

\subsection{Indoor Testing}

In order to compare the output of the developed sensoric system with an absolute and precise measurement, we tested it within an arena, referred as the WhyCon system, developed within our group that uses a downwards-looking camera and a pattern recognition algorithm to compute the position of a ring in a plane \cite{WhyCon}.
For this purpose, we mounted the Raspberry Pi and the pattern on a wheeled carrier and used a circular trajectory and a square one.

\begin{figure}[ht!]
  \centering
  \includegraphics[width=0.7\columnwidth]{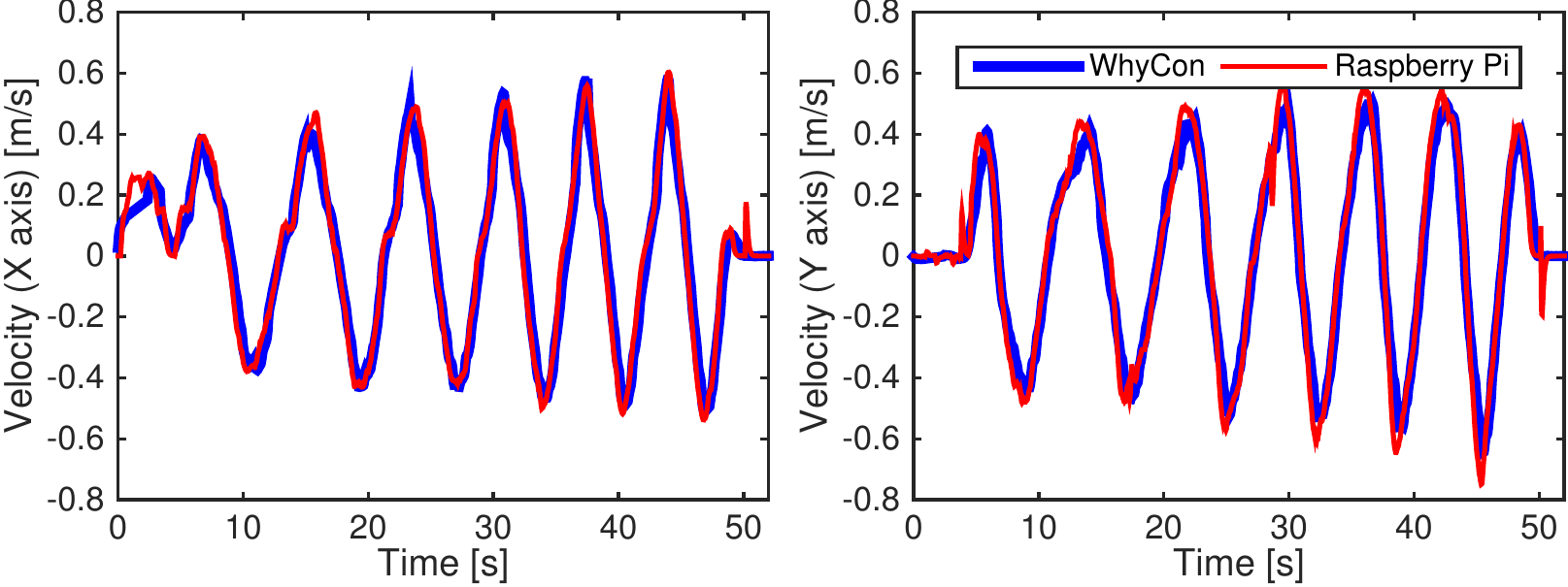}
  \caption{Measured velocity for trajectory ``Circles'' compared to WhyCon}
  \label{fig:ind-vels}
\end{figure}

Figure~\ref{fig:ind-vels} shows the velocities measured by the Raspberry Pi compared to derived and filtered positions measured by the WhyCon system.
Average errors and standard deviations are shown in Table~\ref{tab:ind-vel-errs}.

The processing time is approximately constant and its average value is \unit[5.36]{ms}.
The number of iterations was set to 210 so that the probability of selecting an uncontaminated sample of two motion vectors from the set with up to 85\% outliers is 0.99.
This could be further lowered by using the adaptive method described by Hartley and Zisserman \cite{HartleyZisserman}.

\begin{table}[ht!]
  \centering
  \caption{Measured velocity errors}
  \begin{tabular}{|c|c|c|}
    \hline
    Trajectory & $\mu$ [\unit{m$\cdot s^{-1}$}] & $\sigma$ [\unit{m$\cdot s^{-1}$}] \tabularnewline
    \hline
    \hline
    ``Squares'' & $0.050$ & $0.044$\tabularnewline
    \hline
    ``Circles'' & $0.042$ & $0.036$\tabularnewline
    \hline
  \end{tabular}
  \label{tab:ind-vel-errs}
\end{table}

\subsection{Outdoor Testing}

Outdoor tests have been performed in a park environment using a commercially available hexacopter DJI F550 equipped with a GPS unit and the PX4Flow sensor, remotely controlled by an operator.
The altitude data were obtained from the PX4Flow and integrated in post-processing because the ultrasonic sensor on our system would otherwise interfere with the one of the PX4Flow.
The orientation was ignored during this experiment in order to compare measurements with the PX4Flow sensor which does not recover orientation from the optical flow.

\begin{figure}[ht!]
  \centering
  \includegraphics[width=0.7\columnwidth]{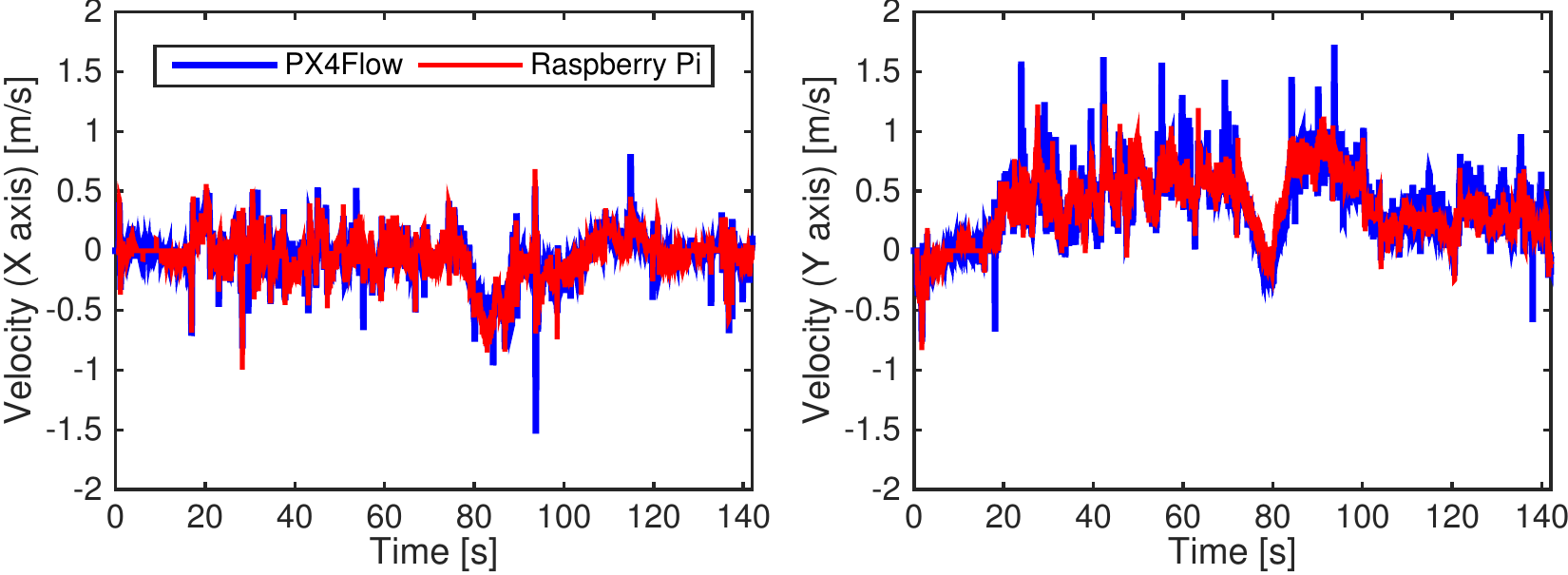}
  \caption{Computed velocity compared to PX4Flow during outdoor experiment}
  \label{fig:Comparison-of-speed-px4flow}
\end{figure}

Fig.~\ref{fig:Comparison-of-speed-px4flow} shows the comparison of velocities between the Raspberry Pi and the PX4Flow.
The average difference is \unit[0.076]{m$\cdot$s${}^{-1}$}.

\begin{figure}[ht!]
  \begin{minipage}[l]{0.45\textwidth}
    \includegraphics[width=1.0\textwidth]{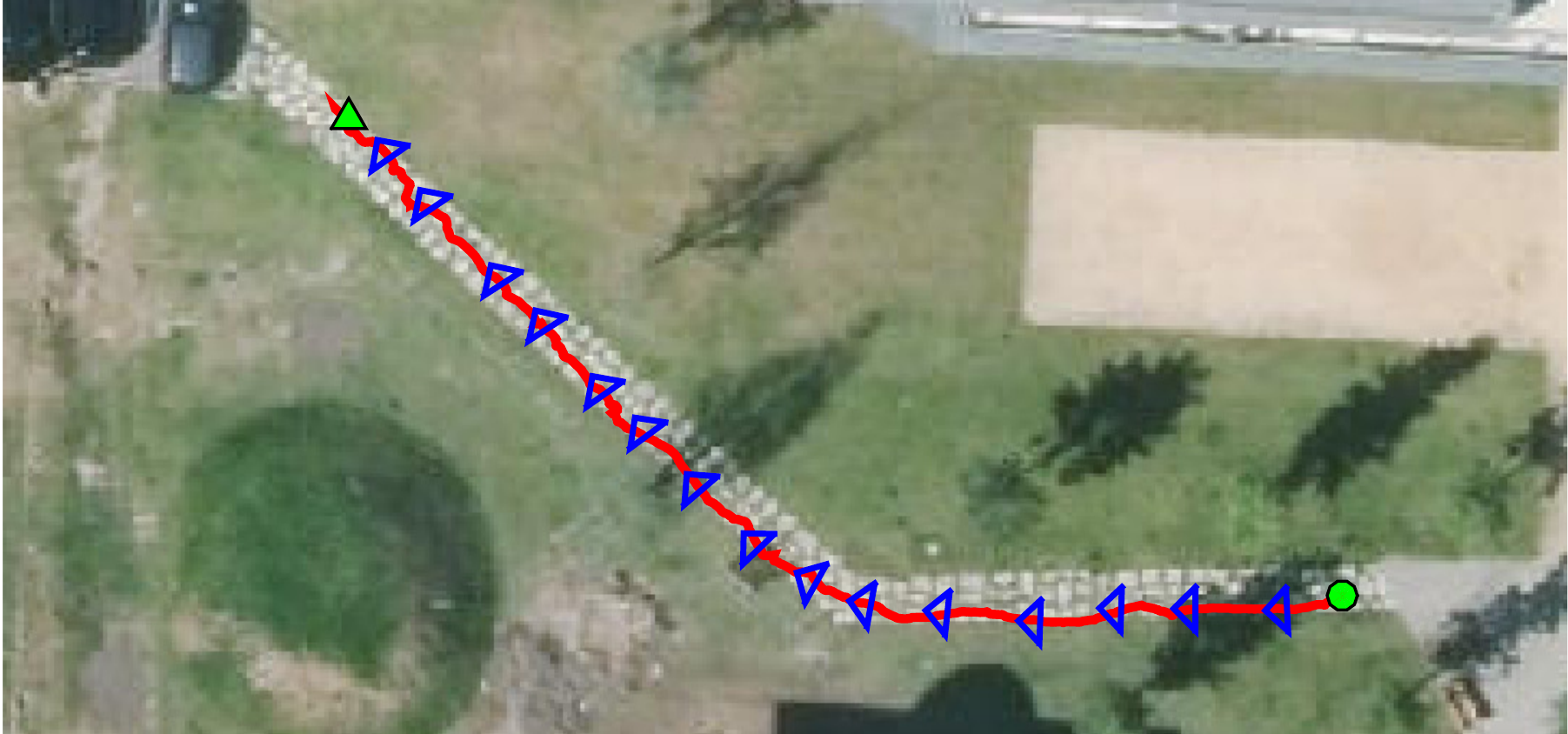}
    \caption{Integrated trajectory compared to map (map from IPR Praha\cite{Map})}
    \label{fig:Integrated-trajectory-compared}
  \end{minipage}% The % symbols avoid a new paragraph!? Leave them!
  \hfill%
  \begin{minipage}[r]{0.45\textwidth}
    \includegraphics[width=1.0\textwidth]{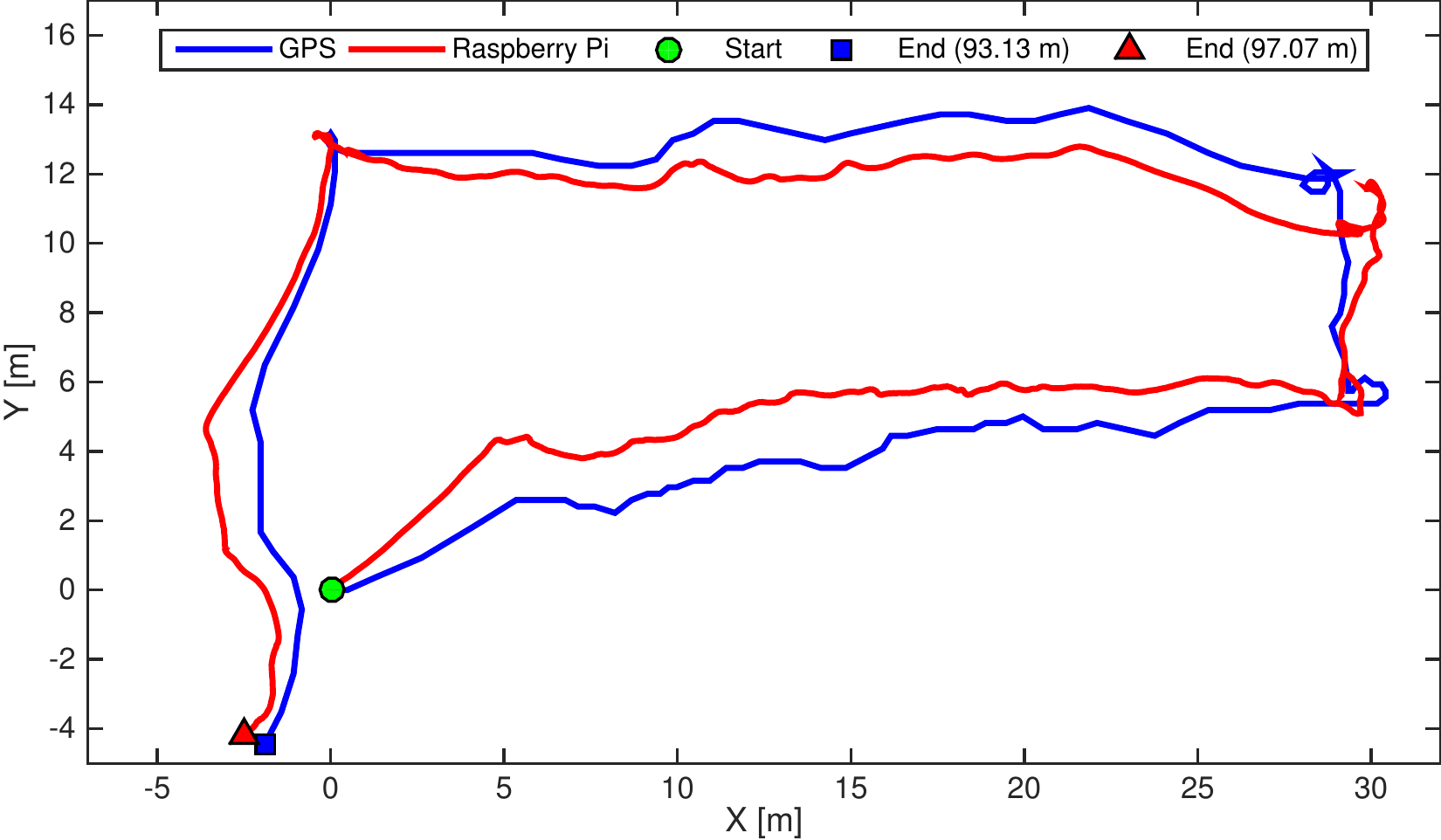}
    \caption{Integrated trajectory compared to GPS}
    \label{fig:Position-compared-to}
  \end{minipage}
\end{figure}

Fig.~\ref{fig:Integrated-trajectory-compared} shows a different trajectory recorded during a flight above pavement to demonstrate the ability to recover changes in orientation.
Fig.~\ref{fig:Position-compared-to} shows the position integrated by the Raspberry Pi compared to the position recorded from the GPS receiver.
The geographic coordinates have been approximately converted to meters using the WGS~84 spheroid \cite{WGS84}, both trajectories have been aligned to have the same origin and, arguably, they have been rotated to have the best fit over the complete trajectory.
The comparison shows that our sensoric system provides coherent data also on the long term.

\section{Conclusion and Perspectives}

We presented an ego-motion sensor based on the Raspberry Pi 3 and other off-the-shelf components.
The accuracy of the system is comparable with the one of the PX4Flow and coherent when compared to some GPS data.
Its average power consumption is \unit[390]{mA} at \unit[5]{V}, compared to \unit[115]{mA} for the PX4Flow.
The advantages of our system over the PX4Flow, however, is that our system is able to provide the changes in orientation around the vertical axis and is cheaper.
Moreover, as the average CPU usage is approximately \unit[23]{\%} at \unit[30]{fps} and the memory footprint is under \unit[30]{MB}, the computing resources of the Raspberry Pi are not saturated and let room for other algorithms.
To foster further developments of the system we plan to publish the source code with an open-source license.

Our further work will be the integration in the control-loop of the MAV and the integration of a ego-motion computation algorithm based on feature detection to improve the low-speed behavior, particularly its inherent drift.
Tests over different surfaces also belong to the plans for future work.

\bibliographystyle{ws-procs9x6}
\bibliography{references}

\end{document}